\documentclass[conference]{IEEEtran}
\IEEEoverridecommandlockouts
\usepackage{cite}
\usepackage{amsmath,amssymb,amsfonts}
\usepackage{algorithmic}
\usepackage{graphicx}
\usepackage{textcomp}
\usepackage{xcolor}
\usepackage{nccmath}
\usepackage{booktabs}
\usepackage{relsize}
\usepackage{url}
\usepackage{makecell}
\usepackage{nicefrac}
\usepackage{todonotes}
\usepackage{multirow}
\usepackage{amsthm}

\newcommand{\norm}[1]{\left\lVert#1\right\rVert}

\theoremstyle{definition}
\newtheorem{definition}{Definition}

\def\BibTeX{{\rm B\kern-.05em{\sc i\kern-.025em b}\kern-.08em
    T\kern-.1667em\lower.7ex\hbox{E}\kern-.125emX}}

\begin{document}

\title{Hierarchical Target-Attentive Diagnosis Prediction in Heterogeneous Information Networks}



\author{\IEEEauthorblockN{Anahita Hosseini}
\IEEEauthorblockA{\textit{Department of Computer Science} \\
\textit{University of California, Los Angeles}\\
Los Angeles, USA \\
anahosseini@cs.ucla.edu}
\and
\IEEEauthorblockN{Tyler Davis}
\IEEEauthorblockA{\textit{Department of Computer Science} \\
\textit{University of California, Los Angeles}\\
Los Angeles, USA \\
tylerdavis@cs.ucla.edu}
\\
\and
\IEEEauthorblockN{Majid Sarrafzadeh}
\IEEEauthorblockA{\textit{Department of Computer Science} \\
\textit{University of California, Los Angeles}\\
Los Angeles, USA \\
majid@cs.ucla.edu}
}

\maketitle

\begin{abstract}

We introduce HTAD, a novel model for diagnosis prediction using Electronic Health Records (EHR) represented as Heterogeneous Information Networks. Recent studies on modeling EHR have shown success in automatically learning representations of the clinical records in order to avoid the need for manual feature selection. However, these representations are often learned and aggregated without specificity for the different possible targets being predicted. Our model introduces a target-aware hierarchical attention mechanism that allows it to learn to attend to the most important clinical records when aggregating their representations for prediction of a diagnosis. 

We evaluate our model using a publicly available benchmark dataset and demonstrate that the use of target-aware attention significantly improves performance compared to the current state of the art. Additionally, we propose a method for incorporating non-categorical data into our predictions and demonstrate that this technique leads to further performance improvements. Lastly, we demonstrate that the predictions made by our proposed model are easily interpretable.
\end{abstract}



\section{Introduction}

Electronic Health Records (EHR) provide a comprehensive picture of patients' medical histories, consisting of information such as written clinician notes, medical imagery, prescriptions, and diagnoses. With the recent availability of EHR datasets to researchers, there has been a significant amount of interest in using this information to improve patient outcomes. In this study, we focus on the problem of predicting patients' diagnoses based on their health records.

Some of the challenges in mining health data are its high heterogeneity and its sparse record distribution, which have led many studies to rely on expert knowledge and manual selection of a set of dense features~\cite{harutyunyan2017multitask, purushotham2017benchmark}. One way in which these challenges have been approached is through an unsupervised record embedding technique, first proposed by Med2Vec~\cite{choi2016multi}. Med2Vec, as well as successive studies such as~\cite{choi2017gram}, use a skip-gram~\cite{mikolov2013distributed} based technique to learn latent representations for health records based on their co-occurrence relations. In this approach, predictions are commonly made by training supervised models on patient representations, which are obtained by aggregating the embeddings of the items in a patient's health records.

Another work using a similar approach is HeteroMed~\cite{Hosseini:2018:HHI:3269206.3271805}, which demonstrates the advantages of modeling EHR data using Heterogeneous Information Network (HIN). HeteroMed shows that HINs can capture the structure and semantically important relations of EHR and model its heterogeneity. In this study we continue to explore the promise of HINs for modeling EHR, addressing the shortcomings of prior record embedding approaches along the way.


One shortcoming in these past works stems from the relatively simple aggregation process they use, in which they treat records with equal importance regardless of what diagnosis is being predicted. Taking diabetes and kidney failure as an example, we can see how this is an issue: prior models generate a single patient representation by combining records with fixed weights, which is then used for the prediction of both diagnoses; however, the importance of tests should vary based on the diagnosis being predicted, with blood glucose levels being more important than blood albumin levels when predicting diabetes than when predicting kidney failure and vice versa. 

Another shortcoming of these past approaches is that the predictions generated by these models are not easily interpretable, with no way for an end user to understand how the model arrived at its conclusion. Lastly, past approaches only make use of records whose values can be mapped to distinct categories, leaving out other important information such as time series vital signs and medical imagery.


Inspired by the very recent success of attention mechanisms in network embedding~\cite{zhou2019hahe, velivckovic2017graph}, we propose HTAD, a novel approach for modeling EHR data that leverages hierarchical attention, to overcome these shortcomings. HTAD produces diagnosis-aware patient representations, as well as explainable predictions. We also suggest how non-categorical data, in particular, time series data, can be integrated into HTAD. 

Considering EHR in the context of HIN with patients and records mapped to network nodes, our model's goal is to aggregate a patient's neighborhood such that the obtained representation is tailored to the prediction of a specific target diagnosis. Recognizing heterogeneity of nodes, we perform the neighborhood aggregation at two levels: first, at node-level and among nodes having similar type to obtain a set of type representations, and then at the type-level to achieve a comprehensive patient representation. In node-level aggregation, we propose employing a target-aware attention mechanism to learn the importance of various nodes with respect to the given diagnosis. We also show ways for the incorporation of time-series data at this level. We apply similar attention technique at the type-level to allow the model to learn preference towards various record types for the prediction of the specified disease.  We then pass the resulting patient representation into our objective function for prediction. Importantly, attention weights generated in our model improve the interpretability by providing insight as to which nodes and types the model finds most important for the prediction. 

We evaluate our proposed model's performance on two diagnosis prediction tasks: exact diagnosis code prediction and high-level diagnosis group prediction, using the publicly available MIMIC-III EHR dataset~\cite{johnson2016mimic}. We compare HTAD to several existing models that represent the state of the art for diagnosis prediction using EHRs.
Our experiments show that HTAD outperforms these benchmarked models on both tasks, in multiple cases beating them by a margin of over 10\%.

Additionally, we evaluate our model's interpretability, something that has not been explored in past models for diagnosis prediction that represented patients based on their aggregated EHR embeddings. In summary, we make the following contributions in this paper:

\begin{enumerate}
\item We propose Hierarchical Target Attentive Diagnosis (HTAD) in an HIN setting and demonstrate that it significantly improves diagnosis prediction performance.
\item We demonstrate that HTAD's use of target-aware hierarchical attention can improve interpretability.
\item We demonstrate that non-categorical data can be incorporated when mining EHR data represented as an HIN.
\end{enumerate}


\section{Related Work}
In this section, we highlight prior representative works in three areas that come together in this study: EHR data mining, Heterogeneous Information Network embedding, and attention-based modeling. 

\subsubsection{EHR Modeling}
When modeling EHR, there are two main challenges that prior studies have approached.
First, clinical records are heterogeneous and are sparsely distributed among patients. To tackle this, manual feature selection has been a method of choice in many studies, leading to two recent works on benchmarking a public EHR dataset~\cite{harutyunyan2017multitask, purushotham2017benchmark} and introducing a set of features to be extracted for various tasks~\cite{harutyunyan2017multitask}. In another direction, studies such as Med2Vec~\cite{choi2016multi} introduced the unsupervised embedding of clinical records using a skip-gram which was adopted by a number of later studies~\cite{choi2017gram, ma2017dipole, choi2016retain} and was extended by Heteromed~\cite{Hosseini:2018:HHI:3269206.3271805}.

Second, it can be difficult to model the complex structure and relations in EHRs. Recurrent Neural Networks (RNNs) have been one of the most widely adopted techniques. However, RNNs lose efficiency and performance when working on long sequences, and clinical records may contain thousands of items. Moreover, they fail to capture the structure and semantics of relations in EHR. HeteroMed~\cite{Hosseini:2018:HHI:3269206.3271805} proposes the use of HINs for the analysis of EHRs, allowing to capture both node and relation semantics. Our work is inspired by the success of HeteroMed in representing EHRs as an HIN and works to overcome prior studies shortcomings in disregarding the importance of records and providing integrative modeling.

\subsubsection{Heterogeneous Information Network Embedding}
\begin{figure}
    \centering
    \includegraphics[width=\linewidth]{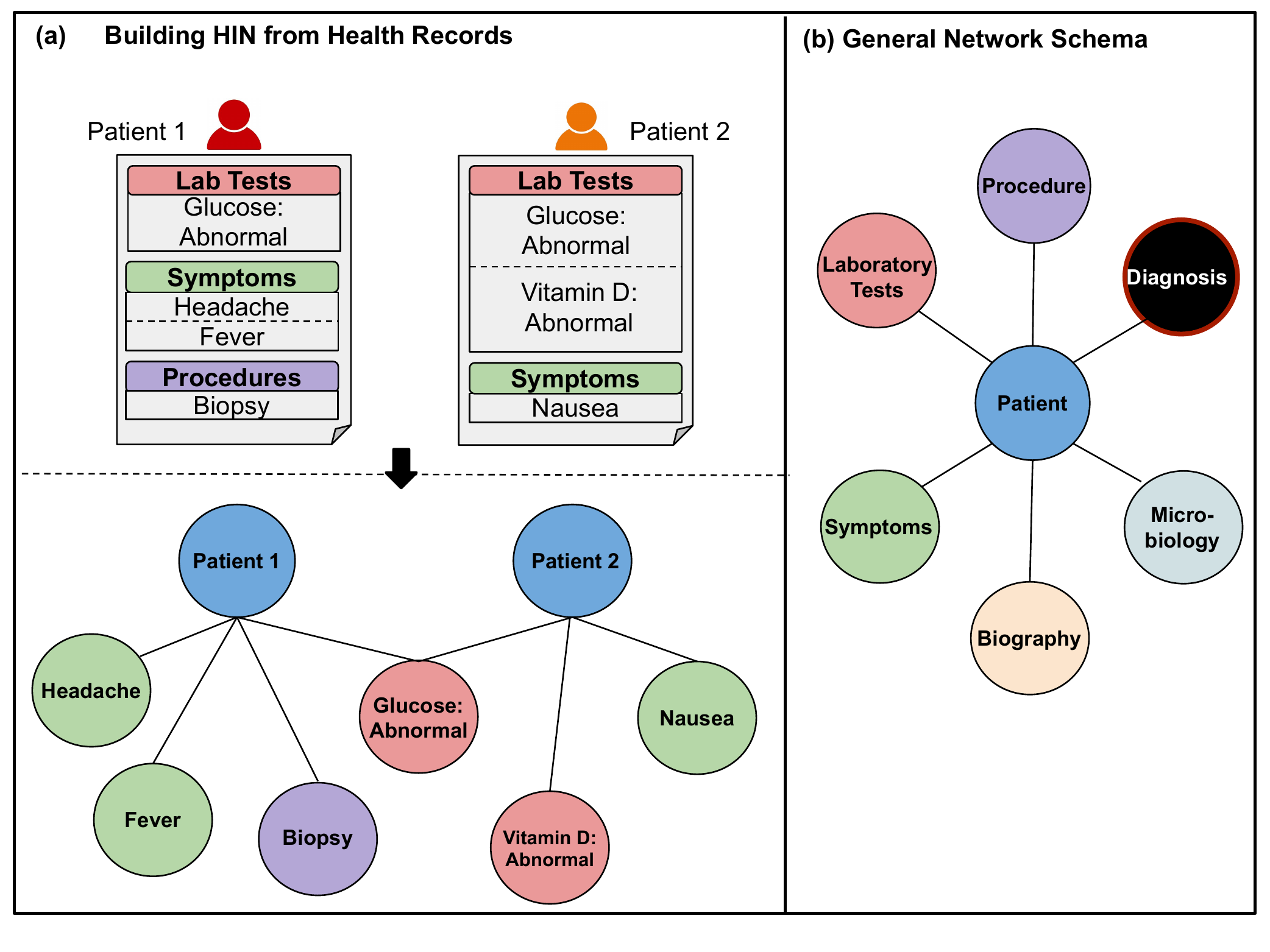}
    \caption{(a) A visualization of how we map EHRs to an HIN, (b) EHR heterogeneous information network schema.}
    \label{fig:subject}
\end{figure}
Heterogeneous Information Networks (HIN) have recently gained considerable attention, especially in the domain of recommendation systems. These networks are able to capture various types of entities and relation semantics, which is essential in modeling real-world settings. Embedding an information network refers to learning compact representation vectors for its nodes. Many homogeneous network embedding approaches, such as DeepWalk~\cite{perozzi2014deepwalk} and node2vec~\cite{grover2016node2vec}, employ random walks or neighbor prediction mechanisms, paired with skip-gram based models. For HINs, relation-based walks have been introduced to incorporate the heterogeneity of data~\cite{dong2017metapath2vec}.
\subsubsection{Attention Mechanisms}
Attention mechanisms for learning algorithms have gained huge success in the domains of natural language processing~\cite{bahdanau2014neural}, with the goal of allowing a model to attend to the most important parts of text while ignoring less relevant portions. Attention for network analysis is a growing topic of interest, with recent studies~\cite{velivckovic2017graph, zhou2019hahe} employing it in the selection of important neighbor nodes, random walks, and meta paths, respectively. In this study, we explore attention in HINs for target-aware node importance scoring when modeling EHR.


\section{Preliminaries}
\label{sec:prelims}
\begin{definition}{\textbf{Heterogeneous Information Networks~\cite{sun2012mining}}}
\textit{A Heterogeneous Information Network (HIN) is defined as a graph $G = (V, E)$ with two type functions $h :V\mapsto A$ and $g: E \mapsto R$ that map nodes and edges to their predefined types $A$ and $R$, respectively.}
\end{definition}
\begin{definition}{\textbf{Meta Path~\cite{sun2012mining}}}
\textit{
Given $A$ and $R$, representing sets of all node and edge types in graph $G$, a meta path is defined by a schema in the form of $A_{1} \xrightarrow{R_{1}} A_{2} \xrightarrow{R_{2}} \ldots \xrightarrow{R_{m}} A_{m+1} $. Any two nodes with a connecting path matching this schema will be linked through this meta path.}
\end{definition}
\subsection{EHR Network Formation Process}
In general, an EHR can be viewed as a set of patients $P=\{p_1, p_2, \ldots, p_{|P|}\}$ and clinical records $C=\{c_1, c_2, \ldots, c_{|C|}\}$. We first put forward a formal view of clinical records.
\begin{definition} {\textbf{Clinical Record}} \textit{
A clinical record is defined as a triple: $c=(i, t, v)$, where $i$, $t$, and $v$ respectively denote the id of the recorded item (e.g., blood glucose level), its type (e.g., laboratory test), and its value which can be null for some record types, such as symptoms.}
 \end{definition}

To model EHR as an HIN we rely on a function mapping clinical records to nodes, defined as: $f_c$: $C \mapsto V$, which projects $c = \{i,t,v\} \in C$ to a node $v \in V$ identified by the tuple $(i,v)$ and having type $t$. Similarly, $f_p:P \mapsto V$ maps each patient to a node with the same type and identified by the patient id. Furthermore, the basic links of the network are formed between patient nodes and the nodes representing their clinical records. Fig.~\ref{fig:subject} illustrates this process. To interpret the clinical record values in an EHR, we follow the strategies introduced in~\cite{Hosseini:2018:HHI:3269206.3271805}, which attempt to categorize all node values. However, unlike their approach, we do not discard information that remains in a non-categorical format and we later present a way for incorporating this data into our model. 
\begin{definition}{\textbf{Target/Context Nodes}} \textit{
Target nodes are defined as the nodes for which the presence of the link to a patient should be predicted (diagnosis nodes in this study). All nodes other than patient and target are considered as context nodes.}
\end{definition}

Given these preliminaries, the diagnosis prediction task in an HIN representing EHR data can be defined as:
\begin{definition}{\textbf{Clinical Prediction in an HIN Setting}} 
\textit{Given a patient $p$ with context nodes $N(p) = \{N_{1}(p), N_{2}(p), \ldots, N_{T}(p)\}$ where $N_{t}(p)$ denotes the type $t$ neighborhood of $p$, predict $p$'s target neighborhood: $N_d(p) = \{d_1, d_2, \ldots, d_{|N_d(p)|} \}$, where $d_{i}$ is the $i$th target node.}
\end{definition}

When working with diagnosis prediction task, it is important to note that many medical ontologies, such as the \mbox{ICD-9} system~\cite{american2004international}, provide a hierarchical and multi-resolution view of diagnoses, with the highest level of the hierarchy identifying the general disease group (e.g., cardiovascular disorders) and lower levels providing more specificity as to the exact diagnosis. Importantly, clinicians may assign codes to a patient at any level. Therefore, the diagnosis prediction task can be defined at two levels:
\begin{itemize}
    \item Low-level (exact) code prediction: Due to the large number of diagnosis codes, this task is approached as a ranking problem, with the aim of scoring positively labeled codes higher than others.
    \item High-level (grouped) code prediction: In this task, we aim to predict all diagnosis groups associated with a patient, formulated as a multi-label classification task. 
\end{itemize}

\section{Methodology}
In this section, we present our proposed HIN-based EHR model, leveraging a hierarchical target-attentive architecture. 
\begin{table}
 \renewcommand{\arraystretch}{1.2}

	\centering
	\caption{Notation and Explanations}
	\begin{tabular}{cl}
		\toprule
		Symbol & Explanation \\
		\midrule
		$h_n$ & Embedding of node $n$\\ \hline
		$h^{'}_{n}$ & Transformed embedding of node $n$\\ \hline
		$N_t(p)$ & Type $t$ neighborhood of patient $p$\\ \hline
		$z_{p,d} ^t$ & \makecell[l]{Aggregated embedding of nodes in $N_t(p)$ \\ with respect to diagnosis $d$}\\ \hline
		$q^d$ & Node-level attention vector for diagnosis $d$\\ \hline
		$s^d$ & Type-level attention vector for diagnosis $d$\\ \hline
		$\alpha_{n,d}^{t}$ & \makecell[l]{Node-level attention score assigned to node $n \in N_t(p)$ \\ when predicting for diagnosis $d$} \\ \hline
		$\beta_{p,d}^{t}$ & \makecell[l]{Type-level attention score assigned to type $t$ representation\\ of patient $p$, when predicting for diagnosis $d$} \\ \hline
		$f_{p,d}$ & Aggregated patient $p$ embedding with respect to diagnosis $d$ \\ \hline
		$M$ & Node embedding lookup matrix\\ \hline
		$Q$ & Node-level attention lookup matrix\\ \hline
		$S$ & Type-level attention lookup matrix \\ \hline
		$W_c^t, b_c^t$ & Transformation parameters for context nodes with type $t$\\ \hline
		$W_d, b_d$ & Transformation parameters for target (diagnosis) nodes\\ \hline
		$W_q, b_q$ & Transformation parameters to obtain node-level attention\\ \hline
		$W_s, b_s$ & Transformation parameters to obtain type-level attention\\ \hline
		$W_t, b_t$ & Transformation parameters for time series type embedding\\ 
		\bottomrule
	\end{tabular}
	\label{tab:notation}
\end{table}

\subsection{Model Overview}
\label{txt:model-overview}
\begin{figure}
    \centering
    \includegraphics[width=\linewidth]{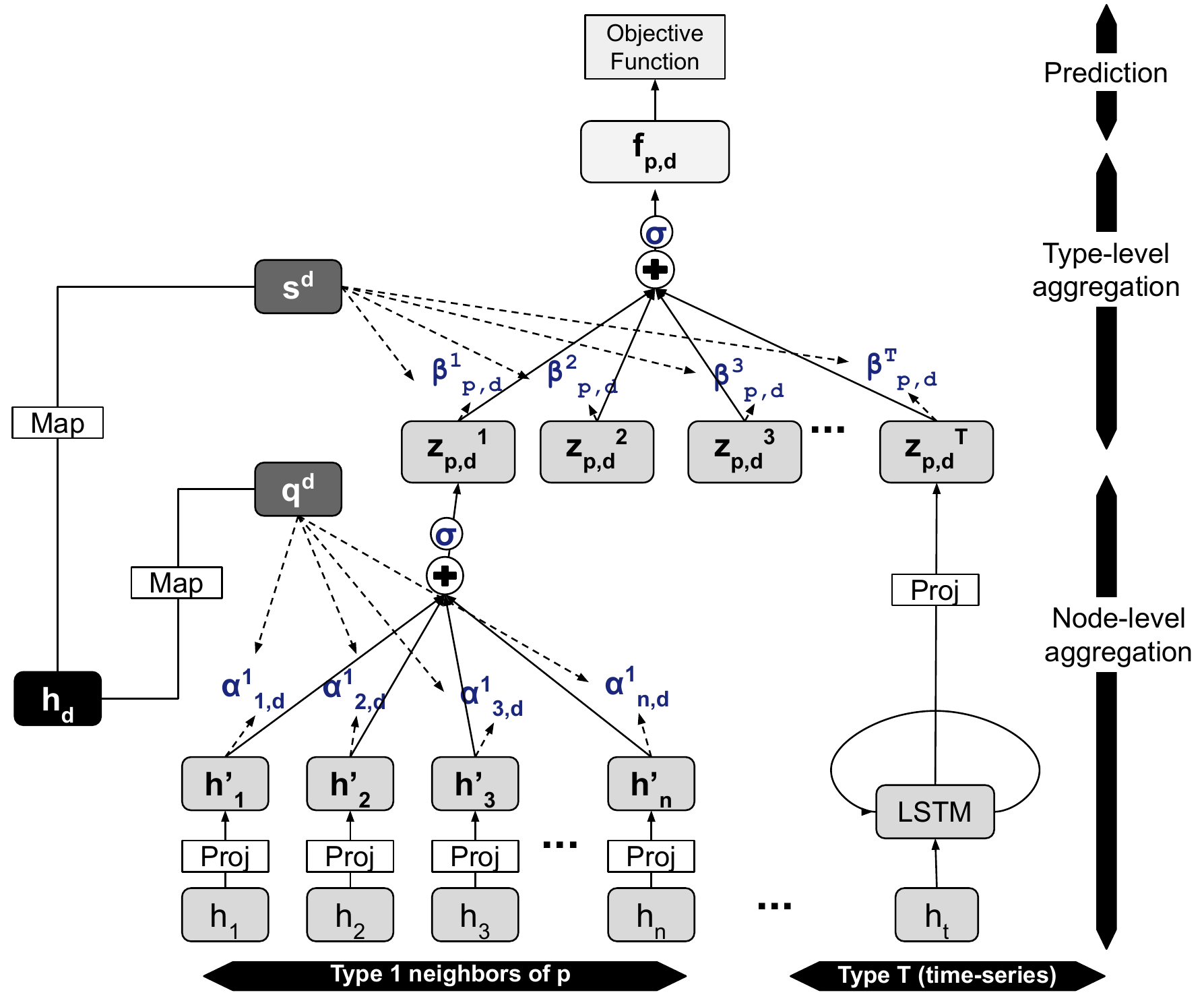}
    \caption{The architecture of the proposed hierarchical target-attentive HIN, illustrating the aggregation of patient $p$'s context nodes with respect to diagnosis $d$. }
    \label{fig:model}
\end{figure}

To model health records and patients, we rely on learning embedding vectors for all these entities. In this approach, a patient representation is often obtained by an aggregation of the embeddings of his/her clinical records and is used for the target prediction task. Different from prior studies where a single patient representation was generated, our model learns to obtain a distinct patient representation for each target node, achieved by favoring the most predictive records for that specific target. The overall architecture for our target-attentive patient aggregation is depicted in Fig.~\ref{fig:model}.

Describing the process in HIN setting, we first aggregate context nodes based on their type using a node-level attention mechanism, generating type-specific embedding vectors. The attention weights are assigned based on the importance of the node in the prediction of the diagnosis. We also present a type-level attention layer to learn the importance of each type in predicting the target, further helping to obtain a diagnosis-aware patient representation. Finally, to generate the aggregated type embedding for time-series nodes as well, we replace the node-level attention mechanism with a deep sequential model.

In addition to learning node embeddings using the supervised model described above, we use an unsupervised approach for learning embeddings in order to capture the structure and semantically important relations in EHRs. 

\subsection{Network Node Embedding}
\label{sec:node_emb}
Having $N$ as the set of all network nodes, the embedding of $n \in N$ is denoted as $h_n$ and is obtained by looking up the corresponding vector from a trainable embedding matrix
 $M \in \mathbb{R}^{|N| \times F}$, where $F$ is the length of the embedding vector.

\subsection{Target-attentive Node-Level Aggregation}
\label{sec:node_attention}
As EHRs are composed of data of heterogeneous types, each node type can carry specific semantic and diagnostic information. Therefore, we start the aggregation process of a patient's neighborhood by combining the context nodes based on their types, thus obtaining type representation vectors. With this in mind, given a patient $p$, its type $t$ neighborhood, $N_{t}(p)$, and a diagnosis node $d$ with corresponding embedding vector $h_d$, the node level target-attention works as follows:

We first utilize a linear transformation layer, parameterized by a type-specific weight matrix $W_c^t \in \mathbb{R}^{F^{'} \times F}$ and bias vector $b_c^t \in \mathbb F^{'}$, to project $p$'s context nodes into a new feature space that is more expressive for attention-based node scoring:
\begin{ceqn}
\begin{equation}
h_n^{'} = W_c^t h_n + b_c^t
\end{equation}
\end{ceqn}
where $h_n$ and $h_{n}^{'}$, having length $F$ and $F^{'}$, denote the original and transformed embeddings of context node $n \in N_{t}(P)$ .

The importance of each node is then measured based on the similarity of its transformed embedding to a diagnosis-specific attention vector $q^{d} \in  F^{'}$. In the most general design, this vector is obtained by applying a linear transformation, parameterized by weight $W_{q} \in \mathbb{R}^{F^{'} \times F}$ and bias vector $b_{q} \in \mathbb F^{'}$, to the diagnosis node embedding $h_d$, formulated as: 
\begin{ceqn}
\begin{equation}
q^{d} = W_{q} h_d + b_{q}
\end{equation}
\end{ceqn}
where $h_d$ is the original diagnosis node embedding.

However, when working with low-level diagnosis codes, there is a significant imbalance in their frequency in a real-world setting. Therefore, the prior approach may face trouble in learning attention vectors for sparser codes. As such, grouping together those with similar diagnostic processes and allowing them to share attention vectors can improve the expressive power of attention for sparser codes. 

Following this idea and taking $D^{'}$ as the set of such a grouping with size $|D^{'}|$, $q^{d}$ can be looked up from an attention matrix $Q \in \mathbb{R}^{|D^{'}| \times F^{'}}$, after mapping $d$ to one of the $|D^{'}|$ diagnosis groups. $Q$ is randomly initialized and jointly trained by the model.
It is important to note that for high-level diagnosis classification task these groups can be defined the same as diagnosis groups we are predicting for. We refer to this approach for the rest of this paper as \textbf{group-based} attention.

Having the transformed node embedding $h^{'}_n$ and diagnosis attention vector $q^d$ obtained, the importance score between them denoted as $e_{n,d}^{t}$, is calculated as:
\begin{ceqn}
\begin{equation}
e_{n,d}^{t} = \frac{q^{d} \cdot h_{n}^{'}}{\sqrt{F^{'}}}
\end{equation}
\end{ceqn}
where $t$ shows the type of node $n$ and division by $\sqrt{F^{'}}$ is used to scale the score for improved performance, following~\cite{vaswani2017attention}.

We then normalize the node importance scores using a softmax function to obtain the attention coefficient $\alpha_{n,d}^{t}$.

\begin{ceqn}
\begin{equation}
\alpha_{n,d}^{t} = \frac{\exp (e_{n,d}^t)}{\sum_{n^{'} \in N_{t}(P)} \exp(e_{n^{'},d}^t) }
\end{equation}
\end{ceqn}

Lastly, the normalized attention coefficients are used as weights for linear aggregation of transformed node embeddings, which is then followed by a non-linearity function to form the type embedding: 

\begin{ceqn}
\begin{equation}
z_{p,d}^{t} = \sigma \Big(\sum_{n \in N_t(p)} \alpha_{n,d}^{t} \cdot h^{'}_{n} \Big)
\end{equation}
\end{ceqn}
where $z_{p,d}^t$ denotes the representation of type $t$ neighbors of $p$ when predicting for diagnosis $d$.

\subsection{Node-Level Time Series Aggregation}
As discussed in section~\ref{sec:node_emb}, the node embeddings used in the node-level aggregation process are obtained using a shallow embedding lookup process. However, such a technique is not usable for records kept in a time series format, as these records cannot be easily mapped to a small fixed set of categorical values and as there would be too little sharing of nodes between patients if each unique time series value were mapped to a node. Therefore, to incorporate such records into our proposed information network, we employ a Long-Short Term Memory (LSTM)~\cite{hochreiter1997long} sequential model similar to \cite{harutyunyan2017multitask}. In particular, patient $p$'s time series records $S_t(p) = \{s_1, s_2, s_3, \ldots, s_T\}$ is first fed to the LSTM model and then the hidden state of the last LSTM cell, denoted as $v_t$, is transformed to a vector with embedding size $F^{'}$, forming the type $t$ representation:
\begin{equation}
    z_{p,d}^t = W_t v_t + b_t
\end{equation}
It is worth noting that the embedding obtained is not diagnosis specific, but we have included $d$ to keep the type representation notation consistent throughout the paper.

\subsection{Type-level Aggregation}
\label{sec:type-level_agg}
After deriving type representations, $Z_{p,d} = \{z_{p,d}^{1}, z_{p,d}^{2}, \ldots, z_{p,d}^{T} \}$, our next step is to combine them to generate the patient representation. Similar to nodes, the predictive power of the different types may vary across diagnoses. For example, the diagnosis of some diseases relies more upon the laboratory tests while others on symptoms. 

Therefore, we propose to use another layer of diagnosis-aware aggregation. Similar to node-level aggregation, a type-level attention vector is employed that can either be obtained by a linear transformation of the original diagnosis embedding, parameterized by weight $W^s$ and bias $b^s$, or be looked up from the attention-matrix $S \in \mathbb{R}^{ |D^{'}| \times F^{'}}$.

The normalized attention coefficient between the type $t$ representation ($z_{p,d}^{t}$) and attention vector $s^d $ is defined as:

\begin{ceqn}
\begin{equation}
\beta^{t}_{p,d} = \frac{\mathlarger{\exp \frac{s^{d} \cdot z_{p,d}^{t}}
{\sqrt{F^{'}}}}}
{\mathlarger{\mathlarger{\sum_{ z_{p,d}^{'} \in Z_{p,d}}}
\exp \frac{s^{d} \cdot z_{p,d}^{'}}{\sqrt{F^{'}}} }}
\end{equation}
\end{ceqn}

In the final step, the comprehensive patient representation, specific to prediction of diagnosis $d$, is denoted as $f_{p,d}$ and is obtained by combining the type representations as follows:
\begin{ceqn}
\begin{equation}
f_{p,d} = \sigma \Big(\mathlarger{\sum_{t \in T }} \beta^{t}_{p,d}\cdot z_{p,d}^t \Big)
\end{equation}
\end{ceqn}

\subsection{Model Inference and Optimization}
In section~\ref{sec:type-level_agg}, we explained how we obtain a set of patient representations $F_p = \{f_{p,d_1}, f_{p,d_2},\ldots, f_{p,d_k}\}$, in order to predict each of the $k$ diagnoses in $D=\{d_1, d_2, \ldots, d_k\}$. In this section, we describe the optimization and inference of the two prediction tasks built on top of these representations.
\subsubsection{High-level Diagnosis Code Classification}
As this task is formulated as a multi-label classification problem, we first feed the representations into a Multi Layer Perceptron (MLP) that maps $F_P \mapsto D$ and is implemented in two layers: the first one shared among all patient representations and the second one specific to each diagnosis group. We then optimize the model by the following loss function:

\begin{equation}
\begin{aligned}
&L = mean (l_1, l_2, \ldots, l_k) \\
&l_i = −y_i \log \sigma (x_i) - (1 - y_i) \log (1 - \sigma(x_i))
\end{aligned}
\end{equation}

where $y_i$ denotes the ground-truth label for diagnosis $d_i$ in patient $p$'s records and $x_i$ is the prediction made by the model.


\subsubsection{Low-level Diagnosis Code Ranking}
As this task is framed as a ranking problem, we rely on score calculation between a patient and diagnoses. In particular, given a patient representation $f_{p,d} \in  \mathbb{R}^ {F^{'}}$ learned with respect to diagnosis $d$, the score of diagnosis $d$ for patient $p$ is defined as the dot product between their representations:
\begin{equation}
score (p,d) = f_{p,d} \cdot h^{'}_d
\end{equation}
where $h^{'}_d $ denotes the transformed diagnosis node embedding parameterized by $W_{d} \in \mathbb{R}^ {F^{'} \times F}$ and bias vector $b_d \in F^{'}$, which is in the same space as $f_{p,d}$.

Using this score definition, we optimize the model using a hinge loss formulated as:
\begin{equation}
\max(0, -score(d,p) + score({\sim} d,p) + \epsilon)
\end{equation}
where ${\sim} d$ is a negative diagnosis sampled for this patient and $\epsilon$ is the hinge margin.

\subsubsection{Unsupervised Node Embedding}
\label{sec:unsup_model}
Besides the guidance of the supervised task, the network structure and relation of nodes can provide additional information that can be embedded in node representations. To capture this information, we employ an unsupervised network embedding objective similar to~\cite{Hosseini:2018:HHI:3269206.3271805}. Formally, given a node $i$ and its random neighbor $j$, we calculate the probability of observing $j$ as a neighbor of $i$, conditioned on the type of the simple or meta path $r$ connecting them, as follows:

\begin{equation}
 P(j|i;r) = \frac{\exp(h_i \cdot h_j)}{\sum_{j^{'} \in Dest(r)} \exp(h_i \cdot h_{j^{'}})}
\end{equation}
where $Dest(r)$ is the set of all nodes that are possible destinations on a path of type $r$ and $h_i$ and $h_j$ are the embedding vectors of nodes $i$ and $j$, respectively. As the above probability becomes expensive to compute in large networks, we instead use negative sampling \cite{mikolov2013distributed} to approximate the probability:
\begin{equation}
 \begin{aligned}
 \log P(j|i;r) \approx {} & \log \sigma (h_i \cdot h_j + b_r) + \\
 & \sum_{l=1}^k \mathbb{E}_{j^{'}\sim P^{r}_{n}(j^{'})} [\log \sigma (-h_i \cdot h_{j^{'}} - b_r)]
 \end{aligned}
\end{equation}
The supervised objectives we introduced, try to learn the node embeddings suitable for the diagnosis prediction task, while the unsupervised model embeds more general knowledge about the relation and proximity of nodes. To combine these two types of models, we follow the joint optimization approach suggested in~\cite{Chen:2017:TPH:3018661.3018735} and define the following objective:

\subsubsection{Combining the Supervised and Unsupervised Models}
The supervised objectives we introduced learn the node embeddings suitable for the diagnosis prediction task, while the unsupervised model embeds more general knowledge about the relationa and proximity of nodes. To combine these two types of models, we follow the joint optimization approach suggested in~\cite{Chen:2017:TPH:3018661.3018735} and define the following objective:
\begin{equation}
\begin{aligned}
\mathbb{L}_{joint} = {} & \omega \mathbb{L}_{unsupervised} + \\
& (1 - \omega) \mathbb{L}_{supervised}  + \lambda{\sum_{i} {\norm {h_i}}^{2}_{2}}
\end{aligned}
\end{equation}
where $\omega \in [0, 1]$ sets the weight used when sampling a model to train at each training step. 

\section {Experiments}
In this section, we provide qualitative and quantitative evaluations of HTAD, demonstrating its superior performance to existing models and its interpretability advantages.

\subsection{Dataset}
All evaluation experiments in this study are conducted using MIMIC-III database~\cite{johnson2016mimic}. For data preparation and preprocessing, we follow the steps introduced a recent study on standardizing and benchmarking this dataset~\cite{harutyunyan2017multitask}. Accordingly, a total of 42,019 unique hospital admissions are included for modeling, 35,725 of which are used for training and 6,294 of which are used for testing. A mean of 11 diagnosis codes are recorded for each admission with 6016 diagnosis codes overall. ~\cite{harutyunyan2017multitask} also introduces a set of manually selected features for model training, which we rely upon in our time series node aggregation process. Furthermore, for the task of high-level diagnosis prediction, we rely on the 25 disease phenotype groups introduced in this study. 

\subsubsection{Evaluation Metrics}
Prediction of high-level disease groups is considered a multi-label classification problem. Accordingly, we follow existing works and employ Micro, Macro, and Weighted AUC-ROC scores to evaluate this task. 

On the other hand, the exact diagnosis code prediction task is considered a ranking problem. Following the common approaches in the evaluation of large-scale ranking tasks~\cite{manning2010introduction}, the ranking is conducted on a list of 100 codes, consisting of the original positive codes and a number of negatively sampled diagnosis codes. We evaluate our performance on this task using the Mean Average Precision at K (MAP@K), where K is set to 4, 6, 8, and 10.

\subsection{Baselines}
We compare our proposed model, HTAD, to recent studies that have achieved state of the art results in diagnosis prediction, including those using manual feature selection as well as those relying on unsupervised EHR embedding. We also evaluate variants of HTAD to demonstrate the effectiveness of each of its components. A comprehensive list of models evaluated is as follows:
\begin{itemize}
    \item Std-LSTM~\cite{harutyunyan2017multitask}: An LSTM-based model for predicting high-level diagnosis groups, introduced as the standard baseline for diagnosis prediction task. 
    
    \item MMDL~\cite{purushotham2017benchmark}: A  multi-modal deep model for diagnosis group prediction that relies on a comprehensive set of hand selected features extracted from categorical and time series records in EHR.
    
    \item SAnD~\cite{song2018attend}: A recent study that employs a self-attention mechanism when modeling the EHR data. This study relies on manual feature extraction as well.
    
    \item Med2Vec~\cite{choi2016multi}: An influential skip-gram based model for embedding health records. As this model is used to learn node embeddings and not for prediction, we employ mean aggregation of the embeddings it learns to represent patients based on their records and rely on supervised prediction methods similar to those used in HTAD.
    
    \item HeteroMed \cite{Hosseini:2018:HHI:3269206.3271805}: An HIN embedding method for modeling EHR data. Comparing to HeteroMed can directly reveal the benefits of learning record importance scores, as its basic architecture is similar to HTAD's.
    
    \item HeteroMed\textsubscript{MLP}: A variant of HeteroMed that we use for the group-based diagnosis classification task, obtained by replacing the hinge loss objective with HTAD's multi-label classification one, to achieve a fair comparison.
    
    \item HTAD\textsubscript{noAttnGrp/noTS}: A variant of HTAD that does not employ the group-based attention introduced in section~\ref{sec:node_attention}. This model also excludes time series data so that the performance comparison to HeteroMed is solely focused on the attention mechanism used.
    
    \item HTAD\textsubscript{AttnGrp/noTS}: A variant of HTAD that employs the group-based attention. For fair comparison with HeteroMed, this model excludes the time series data as well. 
    
    \item HTAD: Our proposed model, employing group-based attention along with time series node aggregation.
\end{itemize}

\subsubsection{Implementation Details}
We implemented HTAD in Python using TensorFlow~\cite{abadi2016tensorflow}. HTAD is trained using the Adam optimizer~\cite{kingma2014adam} and the learning rate of the optimizer, the batch size, the node embedding size, and the attention vector size are set to 0.001 and 32,256, and 128 respectively. When using grouped attention vectors, diagnosis groups are formed based on the CCS hierarchical coding system~\cite{healthcarebeta}. Furthermore, the LSTM model used in node-level time series aggregation is pre-trained using the model configuration proposed by the Std-LSTM model~\cite{harutyunyan2017multitask}.

Our implementation of HeteroMed shares its code base with HTAD, particularly in network formation and unsupervised node embedding training. For a fair comparison, both models use the same set of hyperparameters and meta paths when training the unsupervised node embedding task. The metapaths used are: $labt \leftarrow pati \rightarrow diag$, $diag \leftarrow pati \rightarrow symp$, $labt \leftarrow pati \rightarrow symp$. Furthermore, we observed that running the unsupervised part as a pre-training step provided the best results for low-level prediction in HTAD, and as such for both models we do not employ joint training for this task. However, joint training is employed in all other tasks. Med2Vec is trained with an embedding size of 256, and the MMDL and SAnD models are run using the same parameters and setups suggested in their studies. Experiments were run on one NVIDIA GeForce RTX 2080 Ti GPU and two cores on an Intel Core i9-7920X CPU.
\begin{table}[t!]
	\renewcommand{\arraystretch}{1.5}%
	\centering
	\caption{Phenotype Classification Results}
	\begin{tabular}{lrrr}
		\toprule
		& \multicolumn{3}{c}{AUC-ROC} \\
		\cmidrule(r){2-4}
		Model                        & Micro & Macro & Weighted \\  
		\midrule
		Std-LSTM                     & 0.821 & 0.77  & 0.757    \\ 
		MMDL                         & 0.819 & 0.754 & 0.738    \\ 
		SAnD                         & 0.816 & 0.766 & 0.754    \\ 
		Med2Vec                    & 0.815 & 0.748 & 0.741    \\
		HeteroMed                    & 0.831  & 0.745  & 0.739     \\
		HeteroMed\textsubscript{MLP} & 0.864 & 0.788 & 0.786    \\ 
		HTAD\textsubscript{noAttnGrp/noTS}    & 0.871 & 0.829 & 0.815    \\ 
		HTAD\textsubscript{AttnGrp/noTS}    & 0.874 & 0.832 & 0.818    \\ 
		HTAD                         & \textbf{0.880} & \textbf{0.843} & \textbf{0.828}        \\
		\bottomrule
	\end{tabular}
	\label{tab:pheno-res}
\end{table}
\begin{table}
	\renewcommand{\arraystretch}{1.5}%
	\centering
	\caption{Exact Diagnosis Code Ranking}
	\begin{tabular}{lrrrr}
		\toprule
		Model                         & MAP@4 & MAP@6 & MAP@8  & MAP@10 \\ \hline
		Med2Vec                    & 0.752   & 0.743  & 0.738  & 0.714  \\ 
		HeteroMed                     & 0.866   & 0.843  & 0.814 & 0.805  \\ 
		HTAD\textsubscript{noAttnGrp/noTS} & 0.867   & 0.842  & 0.813 & 0.806  \\ 
		HTAD\textsubscript{AttnGrp/noTS}    & 0.888   & 0.848  & 0.821  & 0.810   \\ 
		HTAD                          & \textbf{0.890}    & \textbf{0.881}  & \textbf{0.865}  & \textbf{0.923} \\ 
		\bottomrule
	\end{tabular}
	\label{tab:low-level-res}
\end{table}
\subsection{Evaluation of Disease Phenotype Classification}
Table~\ref{tab:pheno-res} lists the results obtained from evaluating our models on the diagnosis group classification task. Overall, we observe that HTAD outperforms all of the baselines we investigated. Inspection of results further demonstrates that:
\begin{itemize}
    \item HTAD\textsubscript{AttnGrp/noTS} shows notably higher performance than HeteroMed\textsubscript{MLP}. This comparison is important as it demonstrates the effectiveness of our target-attentive aggregation mechanism versus models that otherwise share the same structure.
    \item  Compared to HTAD\textsubscript{noAttnGrp/noTS}, HTAD\textsubscript{AttnGrp/noTS} shows slightly better performance. This indicates that defining independent attention vectors as in group-based attention can be easier to train 
    even when we are working with limited set of diagnoses.
    \item HTAD shows better performance than HTAD\textsubscript{AttnGrp/noTS}, which is expected as the latter does not utilize the time series information in our dataset.
    \item HeteroMed\textsubscript{MLP} outperforms HeteroMed by a considerable margin. This is in line with our expectations, as the original ranking objective used in HeteroMed may not be optimal for multi-label classification, and we expected that adjusting that could improve the performance.
    \item HeteroMed\textsubscript{MLP} shows performance distinctly superior to that of the methods that rely on deep neural networks (SAnD, Std-LSTM, MMDL). This can be attributed to the fact that information networks eliminate the need for manual feature selection and allow for the incorporation of all clinical records.  HeteroMed\textsubscript{MLP} also outperforms Med2Vec, which is expected as it employs a more semantic-aware node representation learning approach.
\end{itemize}
\subsection{Evaluation of Exact Diagnosis Code Prediction}
The feature extraction based studies introduced for evaluation of the previous task have not approached the task of exact disease code prediction, mainly due to the huge size of the prediction space. In this study, we evaluate variants of our model against HeteroMed and Med2Vec, results of which are presented in Table~\ref{tab:low-level-res} that shows:
\begin{itemize}
    \item HTAD, which incorporates time series data as well as group-based attention, outperforms all other models. 
    \item Similar to the high-level classification task, a comparison between HTAD\textsubscript{AttnGrp/noTS} and HeteroMed reveals the significance of employing hierarchical attention mechanism in node-aggregation.
    \item The performance gain of HTAD\textsubscript{AttnGrp/noTS} compared to  HTAD\textsubscript{noAttnGrp/noTS} is significantly more greater in this task. This gain can better demonstrate the advantage of using the group-based attention mechanism. As discussed before, sharing attention vectors among similar diagnoses can result in better performance for less common ones that otherwise remain under-trained.
\end{itemize}
\begin{figure}[t!]
    \centering
    \includegraphics[width=\linewidth]{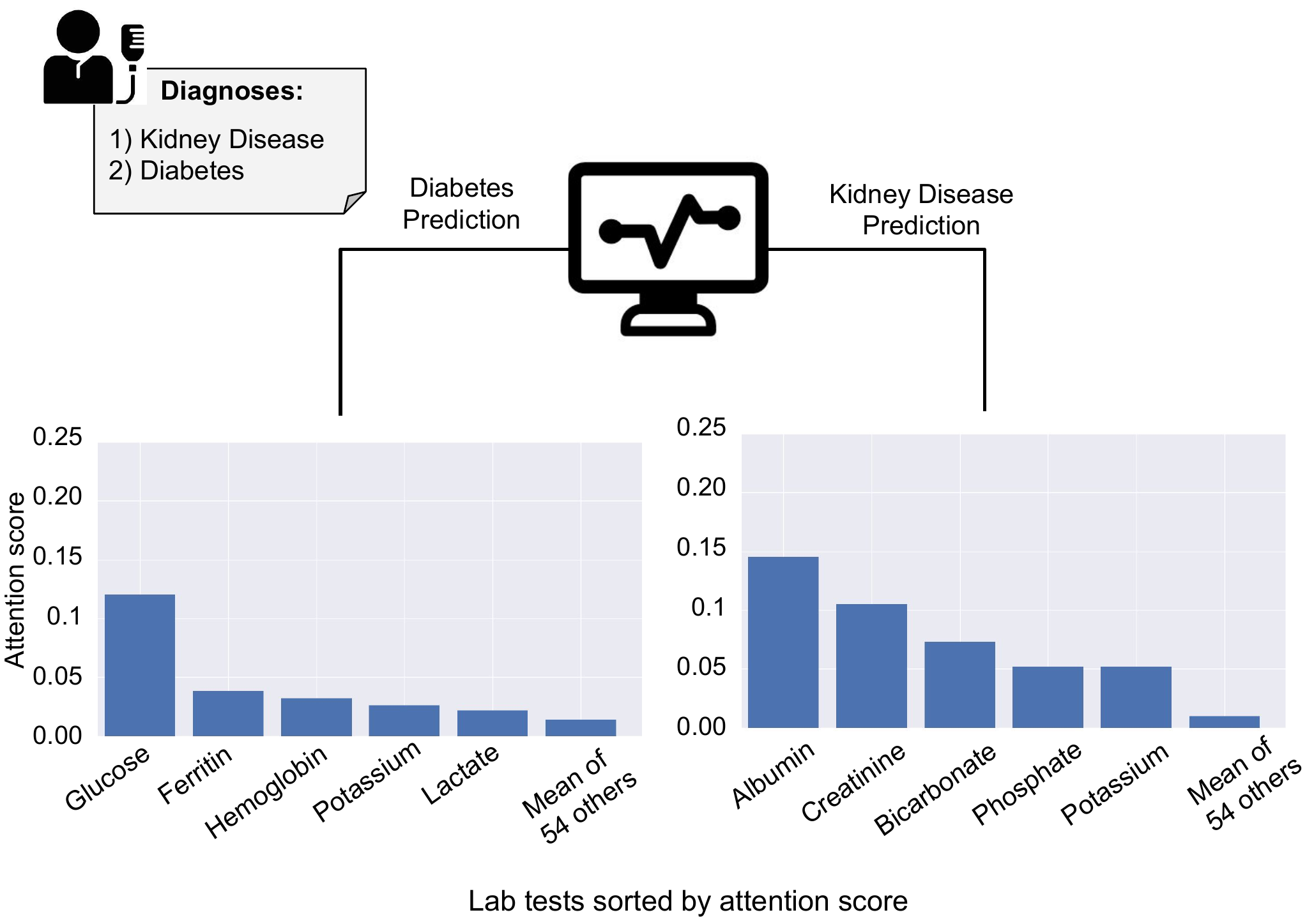}
    \caption{Distribution of attention scores for prediction of kidney disease and diabetes in a patient presenting with both conditions.}
    \label{fig:interpret}
\end{figure}

\subsection{Analysis of Attention Mechanism}
Besides the performance improvement that our proposed hierarchical attentive architecture offers, one major benefit it provides is the interpretability of its results. We illustrate this in the node-level aggregation process in Fig.~\ref{fig:interpret}. We consider a patient diagnosed with both diabetes and kidney failure and study the importance score assigned to each of his 59 laboratory tests when predicting these two conditions. 

The first important observation from this figure is that the set of laboratory tests the model attends to varies between the two diagnoses. As the figure shows, the highest attention score for the detection of diabetes is given to blood glucose level, which is a key predictor for diabetes. Similarly, the laboratory tests listed for kidney failure are highly indicative of this condition. 

Additionally, we observe a larger skewness in attention scores when predicting for diabetes, with glucose having a notably higher score than other labs, than we do when predicting for kidney disease, where attention scores are more evenly distributed. This can be attributed to the fact that kidney failure is indicated by multiple factors while blood glucose is a single key indicator of diabetes. Insights such as these can be highly beneficial in supporting the diagnosis decision process.

\begin{figure}
    \centering
    \includegraphics[width=0.75\linewidth]{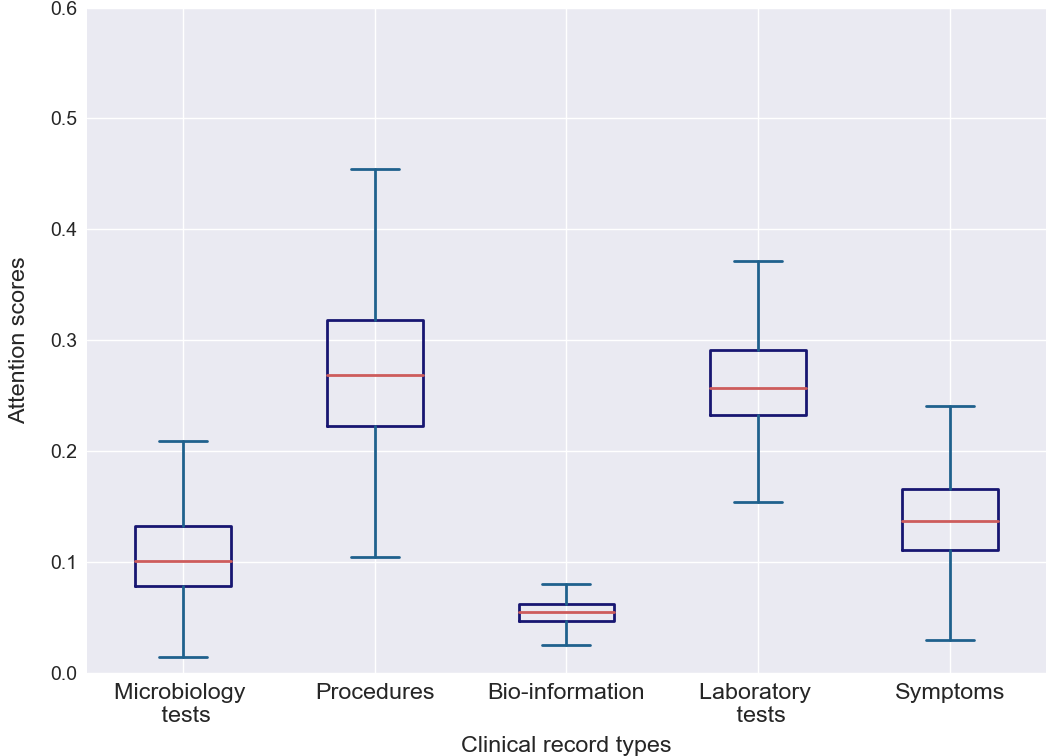}
    \caption{The distribution of attention weights among various types.}
    \label{fig:boxplot}
\end{figure}

We next analyze the attention scores in the type-level aggregation. Fig.~\ref{fig:boxplot} is a box plot demonstrating the range of attention weights assigned to different type-level embeddings across all the diagnoses in our test set. As we can see, the procedures and laboratory tests are overall our main predictors of diagnoses. However, there is more variance in procedure scores than in laboratory test scores, indicating that the predictive power of this category varies across diagnoses.

\section{Conclusion}
In this study, we introduced HTAD, an HIN based model incorporating a hierarchical attention mechanism for diagnosis prediction using EHRs. In HTAD, a patient representation is learned through a target-attentive aggregation of its clinical records' embeddings, a process that allows distinguishing important record items for the prediction of a specific diagnosis. The novelty of this approach lies also in the interpretability it offers. Additionally, HTAD is capable of incorporating non-categorical records unused by past approaches. 
Experimental results demonstrate HTAD's superior performance compared to the previous state of the art methods and the interpretability of its predictions.


\bibliographystyle{unsrt}
\small{\bibliography{refs.bib}}

\end{document}